# Study and Observation of the Variations of Accuracies for Handwritten Digits Recognition with Various Hidden Layers and Epochs using Convolutional Neural Network

Rezoana Bente Arif[1*], Md. Abu Bakr Siddique[1#], Mohammad Mahmudur Rahman Khan[2@], and Mahjabin Rahman Oishe[3$]
[1]Dept. of EEE, International University of Business Agriculture and Technology, Dhaka 1230, Bangladesh
[2]Dept. of ECE, Mississippi State University, Mississippi State, MS 39762, USA
[3]Dept. of CSE, Rajshahi University of Engineering and Technology, Rajshahi 6204, Bangladesh
rezoana@iubat.edu[*], absiddique@iubat.edu[#], mrk303@msstate.edu[@], mahjabinoishe@gmail.com[$]

*Abstract*— Nowadays, deep learning can be employed to a wide ranges of fields including medicine, engineering, etc. In deep learning, Convolutional Neural Network (CNN) is extensively used in the pattern and sequence recognition, video analysis, natural language processing, spam detection, topic categorization, regression analysis, speech recognition, image classification, object detection, segmentation, face recognition, robotics, and control. The benefits associated with its near human level accuracies in large applications lead to the growing acceptance of CNN in recent years. The primary contribution of this paper is to analyze the impact of the pattern of the hidden layers of a CNN over the overall performance of the network. To demonstrate this influence, we applied neural network with different layers on the Modified National Institute of Standards and Technology (MNIST) dataset. Also, is to observe the variations of accuracies of the network for various numbers of hidden layers and epochs and to make comparison and contrast among them. The system is trained utilizing stochastic gradient and backpropagation algorithm and tested with feedforward algorithm.

## I. INTRODUCTION

Deep Convolutional Neural Networks (CNNs) [1-3] have achieved human-level accuracies in many visual recognition tasks including handwritten digits recognition [4, 5], face recognition [6, 7] and traffic sign [4] recognition in recent years. The architecture of the CNN is inspired by the biological modeling of the mammalian visual system. In 1962, D. H. Hubel et al., found that cells in the cat's visual cortex are sensitized to a tiny area of the visual field identified as the receptive field [8]. The neocognitron [9], introduced by Fukushima in 1980 was the first pattern recognition model in computer vision inspired by the work of D. H. Hubel et al. [4, 10]. In 1998, LeCun et al. [11] designed the framework of CNNs with a pioneering seven layered convolutional neural networks [12] adept in classifying handwritten digits directly from the pixel values of the images and could be trained with gradient descent and back propagation algorithm [13]. Since the establishment of the importance of GPGPU for machine learning in 2005 [14], the field of CNN has significantly improved on the best performance in visual classification and extensively uses in several challenging optical recognition tasks of the recent times.

The CNN is very similar in its architecture to a simple artificial neural network (ANN) which is comprised of an input layer, an output layer and some hidden layers in between [15]. In ANN, each layer comprises several neurons. Each neuron in a layer takes outputs from the weighted sum of all the neurons in the precursive layer, and a bias value is added to the result. Then the result is proceeded via an activation function. Unlike ANN, as the neurons in a CNN layer have three dimensions, each neuron in a layer is only connected to a small, localized region of the previous layer known as the local receptive field, instead of all the neuron in a fully connected fashion. As CNN exploits spatial structure in learning and as weights and biases are shared in a receptive field, CNN is much faster than ordinary ANN in operation. To train the network, a cost function is generated to compare the network's output with the desired output to get errors of the results. Then the signal is propagated back to the system in a repeated fashion to update shared weights and biases in all the receptive fields so that the value of cost function is minimized and network's performance is increased [16-18]. Backpropagation algorithm exploits stochastic gradient descent to reduce errors [2, 19].

One of the central mysteries in the field of CNN is the pattern of the implemented hidden layers for the best performance. In some cases, the network converges with a minimal number of hidden layers. Conversely, in some cases, the network needs to have a massive number of hidden layers for the convergence. Therefore, the motivation of this paper is to observe the effects of the hidden layers of a CNN upon the handwritten digits from the MNIST dataset.

This article refers to a model for modeling and simulation of a CNN to recognize handwritten digits from the MNIST database [20]. The mathematical model of this neural network algorithm is implemented in python with numpy and tensorflow. As 28×28 handwritten digits are taken as inputs, this model has 28×28 square of input neurons, in between 5 hidden layers is introduced with two convolution layers, two pooling layers, and one fully connected layer. Output layer is also a fully connected layer which consists of 10 nodes; each of them represents digits from 0, 1, 2, 3, 4, 5, 6, 7, 8, and 9 respectively. Finally, accuracies of the network are observed for a different number of hidden layers, and iterations and comparison were made among them.

## II. LITERATURE REVIEW

In various sectors like image and data processing the CNN has been imposing a substantial impact to perform extensively complex tasks nowadays. From image detection to signal processing CNN is leaving a powerful effect. Moreover, sector like nano-technology is nowadays entirely reliant on CNN which is playing a critical role to detect faults in nanoparticles as well [21]. CNN is also beatitude to the handling of data-sets and large nodes and parameters as well [22]. A heap of researches is befalling for more accuracy and lower loss in CNN. In one prior analysis, varying different parameters of CNN, their effects on CNN was observed. And it was found that the CNN with less number of parameter gives better performance [23]. But in the large-scale neural network, more parameters are involved. Thus it has been an immense challenge to improve CNN performance considering more parameters for better performance. By means of actuating the function of deep-CNN, the noise level in images can be synthesized [24]. Coherence recurrent convolutional network (CRCN) was mentioned to recover compatible sentences in image processing [25]. In image classification, the impact of the parameter of CNN was observed by two datasets; benchmark CIFAR-10 dataset and road-side vegetation dataset. These analyze brought out up to 81% accuracy which is more than the precision found in other datasets such as Alexnet and PSO-CNN [26]. Previously, applying Ncfm (No combination of feature maps) better performance of CNN was illustrated than of the performance in the case of the combination of feature maps using MNIST datasets. In this case, the accuracy was about 99.81% which minimized the problem of dealing with the large-scale neural network [27]. Image reorganization is the crucial sectors of CNN, and its proper performance has been developing day by day through a large number of researches. Error reorganization rates for various epochs were observed both using MNIST datasets as well as using CIFAR10 [28]. Not only in image processing but also in the reduction of noise affect the role of the CNN is essential. A dual-channel model was proposed to have a better performance than of single-channel model [29]. Also, to clean blur images another model using MNIST datasets was proposed which showed the accuracy of 98% and the loss range was 0.1% to 8.5% [30]. Earlier a new model of CNN was suggested for traffic sign recognition in Germany which proposed faster performance to recognize images with 99.65% accuracy [31]. Again, research on loss function calculation was performed using various datasets (CIFAR-10, CIFAR-100, MNIST, and SVHN) [32]. On top of that, loss function for light-weighted 1D and 2D CNN was designed, and for them, the accuracies were about 93% and 91% respectively [33]. In this paper, the performance of the CNN classifier in different cases is tested. The best performance was observed for two consecutive convolutions and pools without dropout.

## III. MODELING OF CONVOLUTIONAL NEURAL NETWORK (CNN)

### A. The Basic Model of a CNN

Like any other standard ANN, a CNN is comprised of an input layer, an output layer and multiple hidden layers in between. The hidden layers of a CNN mainly consist of convolutional layers, pooling layers, and fully connected layers.

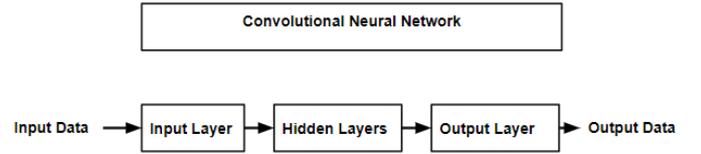

Fig. 1. The basic block diagram of CNN

Though CNN is analogous to traditional multilayer perceptron (MLP) in a sense that both include of neurons that can self-optimize through learning, the notable difference is that CNN exploits three basic ideas absence in traditional ANN [34]: i) Local receptive field, ii) Shared weights, iii) Spatial structure. Figure 2 depicts a typical CNN.

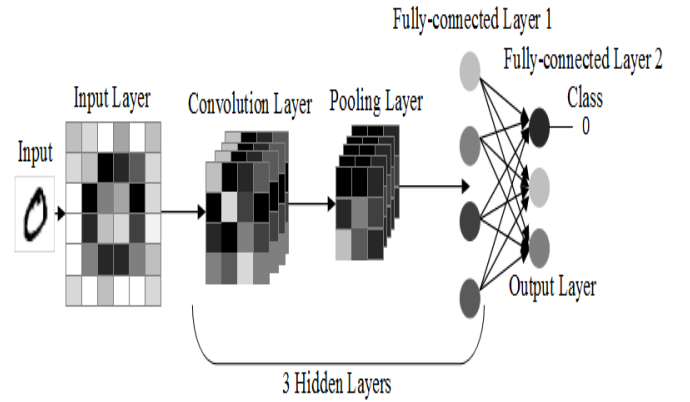

Fig. 2. A typical Convolution Neural Network (CNN)

**Convolution layer:** Convolution layer is the heart of CNN. This layer applies convolution operations to tiny, localized areas of the input feature maps to produce output feature maps. Unlike typical ANN, Each neuron in a hidden layer is linked to a small area of the previous layer identified as the receptive field. Convolution layer consists of learnable filters or kernels with small dimensions. These kernels convolve with receptive fields and spread through the full width, height, and depth of the input volume to produce output feature maps. As the sizes of the kernels define weights, each hidden neuron has a bias and weights same as kernel dimension connected to its receptive field. As equal weights and bias are used for all the neurons in a hidden layer, so technically all the neurons in a hidden layer detect the same feature in different parts of the input layer. That's why map drawn from input to the hidden layer is called a feature map. The weights and bias denoting the feature map are called shared weights and shared bias respectively. The shared weight and bias are used to describe a kernel or filter. Since most of the natural world data are nonlinear, to introduce nonlinearity in CNN an activation function is applied after convolution operations and hence to enhance the performance of the model. The basic equation of convolution operation is expressed in equation 1.

$$\text{Feature map} = \text{input} \otimes \text{kernel} = \sum_{y=0}^{\text{columns}} \left( \sum_{x=0}^{\text{rows}} \text{input}(x-a, y-b)\text{kernel}(x,y) \right)$$
$$= F^{-1}(\sqrt{2\pi} F[\text{input}] F[\text{kernel}]) \quad \ldots\ldots (1)$$

**Pooling layer:** Pooling layer is also an essential unit of a CNN used to additionally cut down the spatial dimensionality of the output of a convolution layer and hence to minimize the number of parameters and computational complexity of the network and to control overfitting. Pooling layer combines the output of neuron clusters in the sub-regions of a convolutional layer into a single neuron in the subsequent layer [3, 35]. The most common types of pooling are max pooling, average pooling [36] and $L_2$ pooling. Max pooling extracts a single maximum value from a cluster of neurons in the precursive layer [4, 37-39]. In recent years, max pooling becomes very popular as it works better in practice compared to other kinds of pooling [40]. As pooling layer dramatically reduces the dimension of the feature map, the trend is to use smaller filters in pooling. In most CNNs, in the case of max pooling, a kernel of 2×2 dimensions with a stride of 2 is used along the spatial dimensions of the input. Stride defines the distance between two consecutive positions of the filter in the input layer along both width and height.

**Fully connected layer:** After several convolution and pooling layers there may have several fully connected layers in CNN architecture. Like any ordinary ANN or MLP, in a fully connected layer, every neuron is attached to all the neurons in the precursive layer. The task of a fully connected layer is to combine features from convolution and pooling layers to produce a probable class score for the classification of the input images. Nonlinear activation function may be used in a fully connected layer to enhance the performance of the network.

*B. A Convolutional Neural Network Model to Classify Handwritten Digits*

To recognize handwritten digits, a seven-layered convolutional neural network with one input layer, one output layer, and five hidden layers is modeled as depicted in figure 3 below.

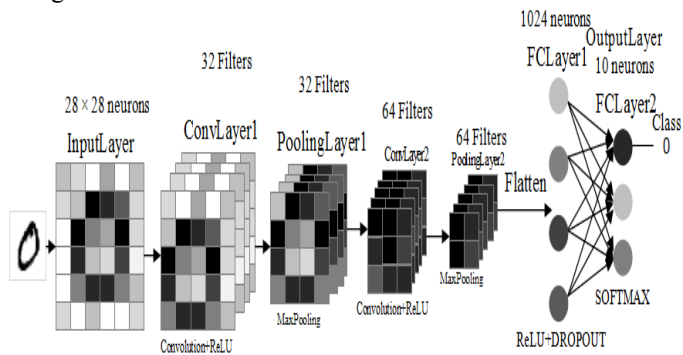

Fig. 3. A seven-layered convolutional neural network for handwritten digits recognition

The input layer of this network contains 784 neurons as the input data consists of 28 by 28-pixel images of scanned handwritten digits. The input pixels are grayscale values with a 0 for a white pixel, and a perfect 1 for a black pixel and different grayscale pixel values are assigned in between zero to one depending on the darkness of the images. This specific CNN has five hidden layers. These layers are convolution layer 1, pooling layer 1, convolution layer 2, pooling layer 2 and fully connected layer one respectively.

To enhance the performance of the model, ReLU is used as an activation function at the end of all convolution layers as well as in fully connected layer 1. In pooling layers 1 and 2, max pooling, a filter of 2×2 dimensions with a stride of 2 is used along the spatial dimensions of the output of the convolution layer to additionally minimize the spatial dimensionality of the output of the convolution layer. In case of convolution layer 1, 32 filters of 5×5 dimensions were used, and for convolution layer 2, 64 kernels of the same sizes were employed. Fully connected layer 1 contains 1024 neurons. To minimize the likelihood to overfit of the network, dropout regularization technique was used at fully connected layer 1 to learn multiple independent representations of the same data by randomly disabling individual activations while training the system. Dropout makes the network more robust to the loss of different pieces of evidence and thus less likely to rely on particular idiosyncrasies of the training data. The fully connected layer 2 or the output layer of the network contains ten neurons to represent digits 0 to 9 respectively. As output neurons are numbered from 0 through 9, the neuron with the highest activation value determines the digit. To enhance the performance of the model, SOFTMAX is used as an activation function at the end of the output layer. Now, a concise explanation of the two activation functions employed in our network is given below.

Rectified Linear Unit (ReLU):
The rectified linear unit [41] outputs 0 if it receives negative input, but if the input is any positive value x, then it returns the same value x. In short, $f(x) = \max(0, x)$. ReLU can also be expressed as shown in equation (2) [42].

$$f(x) = \begin{cases} 0 \text{ for } x < 0 \\ x \text{ for } x \geq 0 \end{cases} \quad \ldots\ldots (2)$$

Softmax Activation Function:
Normally, for classification purpose, the softmax activation function can be employed at the output layer of a CNN. This activation function is better in classification than others as it squeezes the outputs of each segment between 0 and 1. Mathematically, a softmax activation function can be denoted as equation (3) [43].

$$\sigma(z)_j = \frac{e^{z_j}}{\sum_{k=1}^{K} e^{z_k}} \quad \ldots\ldots (3)$$

To train the network 60000 scanned images of handwritten digits leveled with their correct classifications is used from the MNIST database. All photos are grayscale and 28 by 28 pixels in size. After completion of training, the network is then tested with 10000 scanned images of digits. Notation x is used to denote training input. As pictures are 28 by 28 pixels, x is a 784-dimensional vector. The corresponding desired output is denoted by y(x), where y is a 10-dimensional vector.

At the outset of the training, all weights and biases in the network are initialized randomly. As the output of the system solely depends on the shared weight values and the shared bias values of the net, the goal of the network is to find appropriate shared weights and shared biases so that the output of the network approximates desired output y(x) for all training inputs x. To quantify network performances, a cost function is defined by equation 4 [34].

$$C(w,b) = \frac{1}{2n} \sum_x [y(x) - a]^2 \quad \ldots\ldots (4)$$

Here, w = Collection of all shared weights in the network
    b = All the shared biases
    n = Total number of training inputs
    a = Actual output

a depends on x, w, and b. C(w,b) =0, precisely when desired output y(x) is almost equal to the actual output, a, for all training inputs, x. As all parameters in the network are known except b and w, the job of the training algorithm is to find weights and biases so that C(w,b) = 0. So to minimize the cost C(w,b) as a function of the weights and biases, The training algorithm has to find a set of shared weights and biases which make the cost as little as achievable. The algorithm is identified as gradient descent. Gradient descent algorithm utilizes the following equations to set shared weight and bias values to achieve the global minimum of the cost C(w,b) as demonstrated in figure 4. Again, to enhance the performance of the network, the notion of stochastic gradient descent was employed.

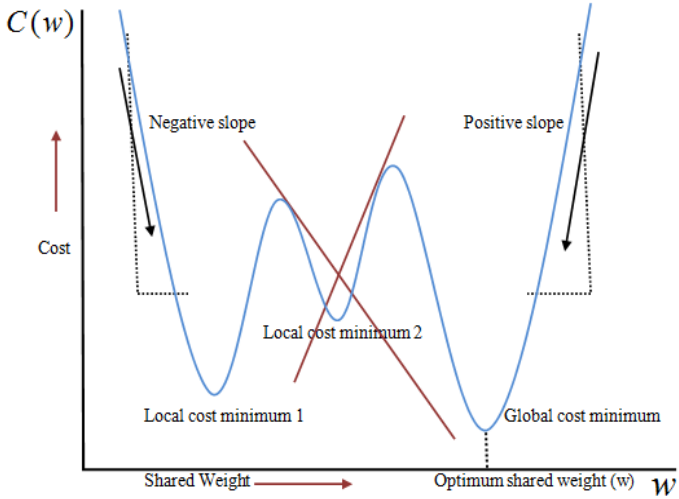

Fig. 4. Cost vs. Shared Weight Plot

Gradient Descent [34]:

$$\Delta C \approx \frac{\partial C}{\partial w} \Delta w + \frac{\partial C}{\partial b} \Delta b \quad \ldots\ldots (5)$$

$$\therefore \nabla C \equiv \left( \frac{\partial C}{\partial w}, \frac{\partial C}{\partial b} \right)^T \quad \ldots\ldots (6)$$

$$w^{new} = w^{old} - \eta \frac{\partial C}{\partial w^{old}} \quad \ldots\ldots (7)$$

$$b^{new} = b^{old} - \eta \frac{\partial C}{\partial b^{old}} \quad \ldots\ldots (8)$$

Stochastic Gradient Descent [34]:

$$C_x = \frac{1}{2}[y(x) - a]^2 \quad \ldots\ldots (9)$$

$$C_x = \frac{1}{n} \sum_x C_x \quad \ldots\ldots (10)$$

$$\therefore \nabla C = \frac{1}{n} \sum_x \nabla C_x \quad \ldots\ldots (11)$$

$$\frac{\sum_{j=1}^{m} \nabla C_{xj}}{m} \approx \frac{\sum_x \nabla C_x}{n} = \nabla C \quad \ldots\ldots (12)$$

$$\nabla C = \frac{1}{m} \sum_{j=1}^{m} \nabla C_{xj} \quad \ldots\ldots (13)$$

$$w^{new} = w^{old} - \frac{\eta}{m} \frac{\partial C_{xj}}{\partial w^{old}} \quad \ldots\ldots (14)$$

$$b^{new} = b^{old} - \frac{\eta}{m} \frac{\partial C_{xj}}{\partial w^{old}} \quad \ldots\ldots (15)$$

Back Propagation:

The output of the network can be expressed by:
$$a = f(z) = f(wa + b) \quad \ldots\ldots (16)$$

Now, the back propagation of the signal can be expressed by the following equations [34]:

$$\delta^{(L)} = \frac{\partial C}{\partial a^{(L)}} \frac{\partial a^{(L)}}{\partial z^{(L)}} = \frac{1}{n}(a^{(L)} - y) f'(z^{(L)}) \quad \ldots\ldots (17)$$

$$\delta^l = \frac{\partial C}{\partial z^l} = \frac{\partial C}{\partial z^{l+1}} \frac{\partial z^{l+1}}{\partial z^l} = \frac{\partial z^{l+1}}{\partial z^l} \delta^{l+1} = w^{l+1} \delta^{l+1} f'(z^l) \quad \ldots\ldots (18)$$

$$\frac{\partial C}{\partial b^l} = \delta^l \quad \ldots\ldots (19)$$

$$\frac{\partial C}{\partial w^{(L)}} = a^{l-1} \delta^l \quad \ldots\ldots (20)$$

IV. RESULTS AND DISCUSSION

For the purpose to find out the better CNN classifier performance, losses for different epoch were observed for various combinations of pool and convolution as depicted in figure 5. In different combination, different responses were found.

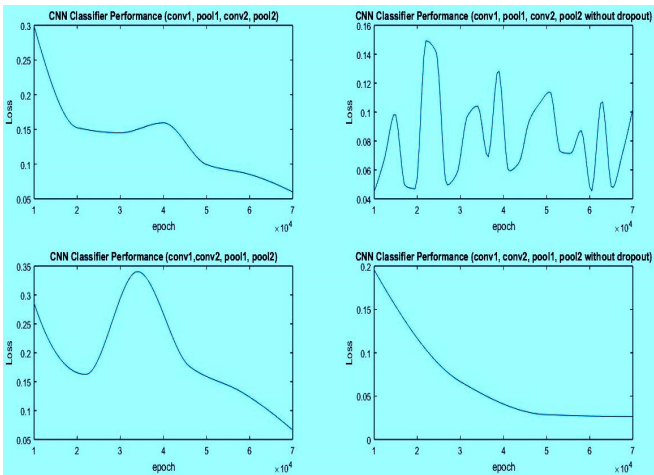

Figure 5: CNN classifier losses for different epoch for various combinations of pool and convolution

For the first case where convolution and pool were in a periodic sequence respectively, the loss vs. epoch responded in a downward direction. For the early $10^4$ epochs the loss was the most, and afterward, it decreased dramatically up to 20000 epochs. Then, the loss stays stabilized approximately at 0.15 up to 30000 epochs. From 30000 to 70000 epochs the loss slowly declined to below 0.1 with a slight increase at 40000 epochs.

If two convolutions were taken first and then two pools were taken the loss curve showed quite identical performance through it gave the highest loss which is more than 0.3 in between 30000 to 40000 epochs. Meanwhile, in the first and fourth combinations, the steepest loss was in 10000 to 20000 range of epoch. And for third observation, the apex was in 20000 to 30000 epochs.

Coming to the third graph in figure 5, where convolution and pool were kept in a repeated sequence without any dropout, the loss fluctuated frantically. In contrast, in the fourth graph with two convolutions and two pools respectively without dropout has demonstrated the best performance of CNN classifier. Here the peak loss was about 0.2 which dropped minimally and became plateau near about 50000 epochs.

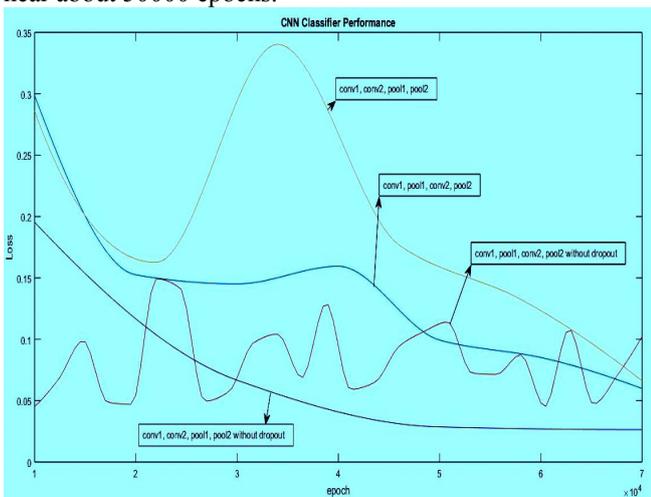

Fig. 6. Comparison of the CNN classifier losses for different epoch for various combinations of pool and convolution

It is evident that the CNN reacts differently in the case of the different combination of layers. Some of the combinations lead to a stable convergence, and others make the convergence process unstable. However, in our work, we found that if the convolutions layers are placed consecutively, the overall performance gets better.

## V. CONCLUSION

In this paper, the loss curves for the separate arrangement of the parameters in the CNN were generated using MNIST datasets. The best and worst both loss curves were found if there was no dropout. The fluctuated and unstable response in the loss was traced if the sets of convolution and pool were one after one without any dropout effect. On the contrary, a smooth degradation in the loss was observed in case of one pair of convolution and one pair of pool respectively. For more than 40000 epochs the loss declined gradually excepting the distorted loss curve (conv1, pool1, conv2, pool2 without dropout). Besides, only in the observation of conv1, pool1, conv2, pool2 without dropout, the loss declined less than 0.04. In all cases, the loss was below 0.1, and in some cases, it was less than 0.05 in unity scale. This proposed low loss response will add up a better performance of CNN to response in noise processing as well as image detection effectively which may attain faster execution of the neural network. In the future, we are going to observe the relation and impact of cost function on CNN performance and better accuracy than the current accuracy in the performance of CNN.